# Effects of Added Emphasis and Pause in Audio Delivery of Health Information

Arif Ahmed, MS[1], Gondy Leroy, Ph.D.[1], Stephen A. Rains, Ph.D.[1], , Philip Harber, M.D.[1], David Kauchak, Ph.D.[2], Prosanta Barai, MS[1]
[1]The University of Arizona, Tucson, Az, U.S.A, [2]Pomona College, Claremont, CA, U.S.A.

**Abstract**

Health literacy is crucial to supporting good health and is a major national goal. Audio delivery of information is becoming more popular for informing oneself. In this study, we evaluate the effect of audio enhancements in the form of information emphasis and pauses with health texts of varying difficulty and we measure health information comprehension and retention. We produced audio snippets from difficult and easy text and conducted the study on Amazon Mechanical Turk (AMT). Our findings suggest that emphasis matters for both information comprehension and retention. When there is no added pause, emphasizing significant information can lower the perceived difficulty for difficult and easy texts. Comprehension is higher (54%) with correctly placed emphasis for the difficult texts compared to not adding emphasis (50%). Adding a pause lowers perceived difficulty and can improve retention but adversely affects information comprehension.

**Introduction**

Increasing health literacy is a major national goal and important for several reasons such as understanding medical information, making informed decisions, managing chronic conditions, preventing health problems, improving health outcomes, and reducing health care costs [1,2]. For many years, written text has been the predominant method for sharing health-related information because it is economical and efficient. With the advancement of technology, novel communication channels are being developed to facilitate access to health-related data, such as interactive videos or audio. Audio is gaining popularity as mobile devices are increasingly being used to listen to information for interaction with virtual assistants and intelligent speakers [3]. Especially the utilization of smart speakers and virtual assistants for health-related inquiries is on the rise. In 2022, it was projected that the United States would have nearly 95 million smart speakers (e.g., Google Assistant, Alexa, Siri) used in households, with an adoption rate of 30-40% each year [3]. These devices are also being integrated into healthcare facilities, enabling patients to engage in conversations with medical professionals and ask them questions [4]. Patients can inquire about health information and listen to the response, via a smart speaker, if the information is straightforward and understandable [5]. In 2019, sixteen percent of all smart speaker inquiries were health-related [6]. The number of American adults using voice assistants for healthcare queries increased rapidly from 19 million in 2019 to 51.3 million in 2020 and 54.4 million in 2021 [7]. Even though there are plain language guidelines for health information delivery using text, audio is overlooked [8]. Integrating audio into current guidelines for disseminating health information could present a significant opportunity to enhance health literacy.

While several methods exist to assess audio or sound quality, no established metric exists to measure the difficulty of information conveyed through audio [9]. Typically, audio is created by recording spoken content and storing it in an audio format. Yet, with modern audio delivery techniques, automated audio generation is emerging as an attractive and feasible alternative. Platforms like Microsoft Azure and Amazon Web Services (AWS) provide tools for selecting between male and female voices, adopting different accents, modifying speech speed, and integrating pauses and emphasis into the produced audio. This study focuses on source text difficulty (easy and difficult) and two audio features (emphasis and pause) and their effect on the perceived and actual difficulty of audio-delivered health information.

## Background

Current studies focus on simplifying text and analyzing text difficulty encompassing various qualitative and quantitative measures. Qualitative factors such as linguistic norms, language structure, clarity, depth of meaning, and the reader's prior knowledge influence the text's perceived difficulty [10]. Features like word frequency, length, and sentence structure influence the quantitative measures. For instance, difficult texts tend to have a lower percentage of verbs and function words and a higher percentage of nouns and use more difficult words (as measured by word frequency) than easy texts [11]. Another study suggests that lexical chains (a text feature) can also be used to discern difficult and easy texts[12]. The variety and spread of topics within a text also matter [12]. Although syntactic and semantic analysis has been the focus of recent studies to reduce text difficulty [13], further study is required to discern the influence of text and audio features on the difficulty of health-related information.

Turning text into audible speech involves translating phonemic elements into sound. To sound genuine to our ears, synthetic speech needs to have the rhythm, inflection, and emphasis that natural speech does. Essentially, this means incorporating prosodic elements. Aspects like timing, pauses, and extended syllables are critical to this. While features such as extending the final syllable have been prevalent in speech synthesis, others, like pauses and rhythm, aren't as common. Fully integrating these timing elements can make synthetic speech sound even smoother, relatable, and easier to understand. [14] For example, a study measuring the frequency and duration of individual pauses during speech found that considering both features is more powerful in measuring the type of decisions such as verbal planning, and generation of information in speech involved in the speech production.[15] Another study on education research shows that the appropriate pause placement during lecture speech enhances both immediate and free-recall performances of upper-level undergraduates in introductory special education courses. [16]

In speech, emphasis serves as a vital cue. It aids in highlighting the central point of what is being said, introducing fresh details, and expressing feelings. [17] Limited research is available on the emphasis and its appropriate placement. One exception is a study on the audio information delivered through an avatar, which showed that adding varying emphasis on the speech can enhance the credibility of the speaking avatars. [18] However, no research is available on the effect of added emphasis on important information delivered over audio. We believe to be the first to evaluate the effects of source text difficulty and audio features such as pause and emphasis in the audio delivery of health information.

## Methods

**Source text.** We generated audio from existing health information texts. To create a set with texts that represent a variety of conditions, we gathered health-related text on diseases from various websites and health-related journals. We created a text corpus from the collected texts and have randomly selected 60 text snippets, representing difficult or easy text (30 text snippets each) for this study. We determined their difficulty based on their source and by referencing the criteria for text complexity we identified in our previous research [11]. The text snippets were about 200-250 words in length. We chose easy texts from the Cochrane Plain Language Summary (15texts) and Rheumatic Disease Journal lay summary (15 texts), and difficult texts were chosen from Wikipedia (5 texts), Rheumatic Disease Journal abstract (6 texts), Medscape (3 texts), and PubMed abstract (16 texts). As intended, there are notable differences between the two groups regarding metrics related to difficulty. For instance, difficult texts had a lower percentage of verbs and a higher percentage of nouns than easy texts. Words are less common in difficult texts than in easy texts (less frequent) [10]. Table 1 outlines the features of these texts. To verify that the conditions differed for the features, we conducted a series of t-tests.

**Audio generation.** We used Microsoft Azure's text-to-speech feature[1] to produce the audio for each text. Azure provides three levels of added emphasis (mild, moderate, and strong) on a selected text. We chose a 'strong' emphasis so that the emphasis was clearly noticeable. We created two versions of each text in which the emphasis was used on information that appears in corresponding multiple-choice (MC) questions for the text (i.e., correct emphasis condition) or information irrelevant to our MC questions (i.e., wrong emphasis condition). We used Azure's pause feature to add a pause before significant information. In generating the audio snippets, we used US male voice and default speech rate.

---

[1] https://azure.microsoft.com/en-us/products/cognitive-services/text-to-speech/

**Study design.** The independent variables of our study are source text difficulty, pause, and emphasis. Text difficulty included a set of more difficult and less difficult texts. The pause variable included no-added pause and added pause prior to the information appearing in the corresponding MC question for the text. The emphasis variable had no-added emphasis, wrongly added emphasis, and correctly added emphasis on significant information. In the condition with both added pause and added emphasis, we added pause before significant information and emphasized the significant information.

Table 1. Text features (T-test, * = p < .05, ** = p < .01, ** = p < .001)

| Variables (Avg) | Easy source text (N=30) | Difficult source text (N=30) | Test Statistic (T-test) | Degrees of freedom | P-value |
|---|---|---|---|---|---|
| **Total characters** | 1368.2 | 1537.233 | -6.7708 | 56.818 | 7.77E-09 |
| **Word counts** | 218.3333 | 226.4333 | -2.5712 | 56.381 | 0.0128 |
| **Sentence length** | 20.02405 | 22.55841 | -2.111 | 44.243 | 0.04046 |
| **Percentage of Noun's** | 30.16667 | 35.93333 | -5.2341 | 57.488 | 2.44E-06 |
| **Percentage of Verb's** | 17.46667 | 13.06667 | 6.0163 | 57.866 | 1.29E-07 |
| **Percentage of Adverb's** | 4.1 | 3.266667 | 1.9045 | 56.93 | 0.0619 |
| **Percentage of Adjective's** | 10.4 | 13.93333 | -4.9737 | 48.635 | 8.59E-06 |
| **Percentage of Function word** | 37.86667 | 33.80000 | 4.6331 | 54.285 | 2.30E-05 |
| **Google word frequency** | 368871654 | 236026454 | 5.5016 | 55.761 | 9.78E-07 |
| **Number of Lexical Chains** | 11.56667 | 13.56667 | -2.2222 | 44.829 | 0.03136 |
| **Chain Length** | 3.287333 | 3.031667 | 2.0966 | 44.361 | 0.04177 |
| **Chain Span** | 102.241 | 109.6773 | -1.0977 | 57.998 | 0.2769 |
| **Number of Cross Chains** | 11.56667 | 13.46667 | -2.1384 | 45.276 | 0.03792 |
| **Number of Half Document Length Chains** | 4.333333 | 5.1 | -1.1929 | 55.127 | 0.238 |

The dependent variables included a perceived difficulty measure and an actual difficulty measure. Study participants were asked to evaluate the perceived difficulty of audio by using a 5-point Likert scale labeled from very easy (1) to very difficult (5). For the actual difficulty measure, we measure information comprehension and retention. To measure comprehension, we formulated multiple-choice (MC) questions using two AI technologies: chatGPT[2] and questgen.ai[3]. Leveraging these AIs, we generated four MC questions for each of the 60 texts. Each question and its corresponding answer was reviewed by a medical domain expert (Dr. Harber) to ensure they centered on the content and had relevant questions with appropriate correct and wrong options. The expert evaluated the generated questions on multiple criteria : question rammar, question relevance, correct answer included, answer grammar, answer semantic type, answer ambiguity, answer comprehension. For information retention, the participants were asked to recall as much information as possible. To analyze the free recall result, we compare two aspects with the original information: The percentage of similar words and the percentage of matching words. The percentage of similar words provides a broader, semantically driven understanding of free recall.

---

[2] https://chat.openai.com/
[3] https://www.questgen.ai/

**Study procedure.** We enlisted study participants through Amazon Mechanical Turk (AMT) AMT assigns workers to participate in each condition. The workers were first asked to complete demographic information about their race, ethnicity, age, level of education, and English-speaking tendency at home. Then, they could complete as many Human Intelligence Tasks (HIT) as they wanted. Each HIT included an audio snippet, followed by four MC questions and a prompt for free recalling the audio information. After listening to the audio clip, the workers responded to questions about its perceived difficulty, answered four MC questions, and provided a free recall response about the audio content they heard.

We sequentially launched the conditions of our study on AMT. Each condition contains 30 audio snippets. Each audio snippet represents one HIT. Based on the condition, the audio is either selected from easy text or difficult text. The workers can participate in multiple HITs. For each HIT, we collected three responses from three different workers. However, the workers were not allowed to work on more than one condition to prevent recalling information from prior participation. Each worker was compensated $1.00 for completing a HIT.

## Results

**Data Cleaning.** For each condition, we collected 90 responses for 30 HITs (3 unique worker's responses for each HIT). We have five conditions, and for each condition, we collected responses for each text difficulty (easy and difficult). That makes 900 responses (5 conditions x 2 text difficulty x 90 responses each) in total.

**Table 2: Example of inappropriate free recall response**

| |
|---|
| "Ponía muy bueno. Eh? La no O no, No me voy a morir. No Yo no puedo más ocupado Mo-. No, no es que no me vaya. It used to sound really good. Huh? No or no, I'm not going to die. No, I can't be busier Mo-. No, it's not that I'm not leaving. Muy bueno. No, no, no, No, no, no me no. No, no voy a vuelto e igual Very good. No, no, no, no, no, not me. No, I'm not coming back anyway". |

We removed data for HITS with inappropriate participation using three principles: we checked the free recall responses to verify the worker's attentive participation; (see Table 2) we checked whether the workers used audio transcription software to answer free recall and no gibberish response by the worker. We also examined the average time it took for each worker to finish a HIT. If the time taken to complete a HIT was shorter than the length of the audio, suggesting they did not listen to the full audio clip, we discarded that data. In addition, we also checked the MC question's responses by each worker. Suppose the average MC accuracy was less than 25% for a particular worker, which is just a chance to be a correct answer for a MC question. In that case, we have removed all the responses of that particular HIT for that worker. As a result, we discarded 223 responses from 53 workers. Ultimately, our data encompasses 677 responses for all conditions.

**Participant demographics.** A total of 172 workers were retained in our study. The majority identified as white (78%) and as not Hispanic or Latino (87%). Slightly more than half were female (53%). Most were 31-40 years (30%) or 41-50 (24%) years old. Most workers had attained a bachelor's degree (51%). And almost all workers (86%) communicated solely in English at home.

**Data Analysis (ANOVA).** Before running the ANOVA, we checked whether responses were independent, normally distributed, and exhibited consistent variance [19, 20]. For the dependent variables, we also produced the Q-Q plot (see Figure 1), which resembles a straight line and depicts the roughly normal distribution of the dependent variables [19]. Additionally, we verified the residual vs. fitted value graph to ascertain consistent variance among groups and found no discernible patterns [20].

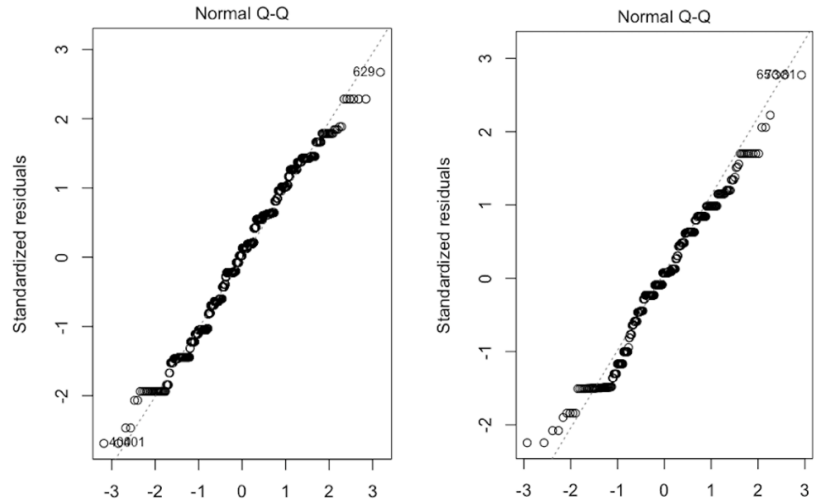

**Figure 1. Normal Q-Q plot for perceived and actual (MC) difficulty responses.**

We conducted two separate ANOVAs for each dependent variable using pause (yes vs. no), emphasis (no vs. correct), and source text difficulty (difficult vs. easy) as independent variables. We conducted separate ANOVAs with the three independent variables because the design was not fully crossed. There is no "wrong" emphasis condition in the added pause section because it did not make sense to have audio with incorrect emphasis and pause before insignificant information.

**Perceived difficulty (Table 3).** For the first ANOVA (text difficulty x pause), we measured the effect of source text difficulty and pause on the perceived difficulty of the audio-delivered health information. We found that there is a significant main effect of source text difficulty ($F_{(1,512)} = 29.84$, $p<0.001$) and audio generated by easier source text resulting in lower perceived difficulty. There was no significant main effect for pause. There is also a significant interaction effect found for source text difficulty and pause ($F_{(1,512)} = 8.70$, $p<0.01$). If we only consider added pause, audio health information delivered with an added pause before significant information is perceived easier (3) than no-added pause for difficult source text (3.5). However, for easy source texts, the added pause did not have any effect on perceived difficulty.

For the second ANOVA (text difficulty x emphasis), we measured the effect of source text difficulty and emphasis on the perceived difficulty of the audio-delivered health information. We found that there is a significant main effect of source text difficulty ($F_{(1,512)} = 30.20$, $p<0.001$). There was a second significant main effect of emphasis ($F_{(1,512)} = 9.38$, $p<0.01$) with correctly added emphasis generates lower perceived difficulty. For the no-added pause section, correctly emphasized significant information is perceived as less difficult (3) than no-added emphasis (3.5) for difficult source text. For easy source text, correctly emphasized significant information is perceived as less difficult (2.5) than no-added emphasis (3). However, wrongly added emphasis on non-significant information perceived is more difficult (4) than any other condition.

**Table 3: Results of Perceived Difficulty. (A lower number indicates easier audio)**

| Conditions | | Source Text Difficulty | |
|---|---|---|---|
| Pause | Emphasis | Difficult (SD) | Easy (SD) |
| No | No | 3.5 (1.1) | 3 (1.2) |
| | Correct | 3 (1.0) | 2.5 (1.0) |
| | Wrong | 4 (1.0) | 3 (1.0) |
| Yes | No | 3 (1.2) | 3 (1.2) |
| | Correct | 3.5 (1.2) | 2.8 (1.0) |

In addition, for the added pause section, correctly emphasized significant information perceived as more difficult (3.5) than no-added emphasis (3) for the difficult source texts. However, for the easy source texts, correctly emphasized significant information is perceived easier (2.8) than no-added emphasis (3). There is a significant interaction effect for source text difficulty and emphasis (F (2,671) =7.059, p<0.05).

**Actual Difficulty**. We analyze information comprehension and retention. For comprehension, we analyzed the MC responses. For retention, we analyzed the free recall responses of the audio information.

*Comprehension (Table 4)*. For the first ANOVA (text difficulty x pause), we measured the effect of source text difficulty and pause on the comprehension of the audio-delivered health information. There is a significant main effect of source text difficulty (F (1,2060) =6.263, p<0.05) with higher comprehension results for audio generated from easy source text. There was no significant main effect for pause. If we only consider added pause before significant information, for the difficult source text, there is no improvement in comprehension result than no-added pause before significant information. There is no significant interaction effect for source text difficulty and pause.

For the second ANOVA (text difficulty x emphasis), we measured the effect of source text difficulty and emphasis on the comprehension of the audio-delivered health information. Natrually, we found again a significant main effect of source text difficulty (F (1,2060) =6.225, p<0.05) but no significant main effect for emphasis. In the no-added pause section, for difficult source text, correctly emphasized significant information yields better accuracy (54%) than no-added emphasis on significant information (50%). However, for the easy source text, added emphasis on significant information did not improve the accuracy.
In the added pause section, correctly emphasized significant information generates lower accuracy (46% for difficult source texts and 49% for easy source texts) than no-added emphasis on significant information. (50% for difficult source texts and 51% for easy source texts). There is no significant interaction effect for source text difficulty and emphasis.

**Table 4: Results of comprehension (%). A higher number indicates better comprehension.**

| Conditions | | Source Text Difficulty | |
|---|---|---|---|
| Pause | Emphasis | Difficult (SD) | Easy (SD) |
| No | No | 50 (41) | 61 (41) |
|  | Correct | 54 (45) | 55 (43) |
|  | Wrong | 52 (38) | 59 (44) |
| Yes | No | 50 (39) | 53 (45) |
|  | Correct | 46 (44) | 52 (44) |

**Retention.** We compare the free recall responses and measure the percentage of exactly matching words and the percentage of similar words.

*Matching words (Table 5)*. For the first ANOVA (text difficulty x pause), we measured the effect of source text difficulty and pause on the retention of the audio-delivered health information. We found there is a significant main effect of source text difficulty (F (1,532) =5.642, p<0.05). There was no significant main effect for pause. Only a correctly added pause before the significant information generates a higher percentage of matching words (13% for the difficult source texts and 11% for the easy source texts) than no-added pause before significant information. (8% for the difficult source texts and 7% for the easy source texts). There is no significant interaction effect for source text difficulty and pause.

For the second ANOVA (text difficulty x emphasis), we measured the effect of source text difficulty and emphasis on the retention of the audio-delivered health information. We found that there is a significant main effect of source text difficulty (F (1,532) =6.432, p<0.05). There is also a significant main effect for emphasis (F (1,532) =65.807, p<0.001) with increased matching words result for correctly added emphasis than no emphasis.

In the no-added pause section, correctly emphasized significant information generates a higher percentage of matching words (22% for difficult source texts and 22% for easy source texts) than no-added emphasis on significant information. (8% for difficult source texts and 7% for easy source texts).

In the added pause section, for the difficult source texts, correctly emphasized significant information generates a lower percentage of 7% of matching words responses than no-added emphasis on significant information for difficult source texts 13%. However, for the easy source texts, correctly emphasized significant information generated a higher percentage of 19% of matching words responses than no-added emphasis on significant information for easy source texts 11%.

**Table 5.** Results of matching words (%). Higher number indicates better retention.

| Conditions | | Source Text Difficulty | |
|---|---|---|---|
| | | Difficult | Easy |
| Pause | Emphasis | Matching Words (SD) | Matching Words (SD) |
| No | No | 8 (5) | 7 (6) |
| | Correct | 22 (11) | 22 (6) |
| | Wrong | 7 (4) | 14 (6) |
| Yes | No | 13 (11) | 11 (8) |
| | Correct | 7 (5) | 19 (7) |

We also found that wrongly emphasized insignificant information generates a lower percentage of matching words responses (7% for the difficult source texts and 14% for the easy source texts) than correctly emphasized the significant information (22% for difficult and easy source texts). There is a significant interaction effect for source text difficulty and emphasis (F (1,532) =12.067, p<0.001).

*Similar words (Table 6).* For the first ANOVA (text difficulty x pause), we measured the effect of source text difficulty and pause on the retention of the audio-delivered health information. We found there is a significant main effect of source text difficulty (F (1,532) =12.130, p<0.001) but no significant main effect for pause. Only correctly added pause before the significant information generates a higher percentage of similar words (16% for the difficult source texts and 15% for the easy source texts) than no-added pause before significant information. (11% for both the difficult and easy source texts). There is no significant interaction effect for source text difficulty and pause.

For the second ANOVA (text difficulty x emphasis), we measured the effect of source text difficulty and emphasis on the retention of the audio-delivered health information. We found that there is a significant main effect of source text difficulty (F (1,532) =13.89, p<0.001) and a significant main effect for emphasis (F (1,532) =71.26, p<0.001).

In the no-added pause section, correctly emphasized significant information generates a higher percentage of similar words (29% for difficult source texts and 30% for easy source texts) than no-added emphasis on significant information. (11% for the difficult source texts and the easy source texts).

In the added pause section, for the difficult source texts, correctly emphasized significant information generates a lower percentage of (9%) of similar words than no-added emphasis on significant information for difficult source texts (16%). However, for the easy source texts, correctly emphasized significant information generates a higher percentage, (25%) of similar words, than no-added emphasis on significant information for easy source texts (15%).

We also found that incorrectly emphasized information generates a lower percentage of similar words (10% for the difficult source texts and 17% for the easy source texts) than correctly emphasized the significant information (29% for the difficult source texts and 30% for the easy source texts). There is a significant interaction effect for source text difficulty and emphasis (F (1,532) =10.91, p<0.01) as well.

**Table 6.** Results of similar words (%). Higher number indicates better retention.

| Conditions | | Source Text Difficulty | |
|---|---|---|---|
| | | Difficult | Easy |
| Pause | Emphasis | Similar Words (SD) | Similar Words (SD) |
| No | No | 11 (5) | 11 (8) |
| | Correct | 29 (13) | 30 (8) |
| | Wrong | 10 (4) | 17 (7) |
| Yes | No | 16 (11) | 15 (9) |
| | Correct | 9 (6) | 25 (8) |

The results of the matching words and the similar words follow the same trends. There is an overall main effect of the pause and emphasis on information retention. Correctly added pause before significant information tends to generate higher retention results than no-added pause before significant information. And correctly emphasized significant information tends to generate higher retention results than wrongly put emphasis or even no emphasis at all.

## Conclusion and Discussion

We explored the effects of two audio features (i.e., emphasis and pause) along with text difficulty on the perceived and actual difficulty of audio-delivered health information. When no pause was added, correctly emphasizing the significant information lowered the perceived difficulty for both easy and difficult source text. Comprehension was also higher on correctly emphasized difficult source texts. An added pause reduced the perceived difficulty more than no-added pause and generated higher retention results. However, for both added pause and added emphasis, the information was perceived as more difficult, and comprehension was also lower.

For information retention, with no-added pause, correctly emphasized significant information generated higher retention results than with no-added emphasis. However, when a pause was included, correctly emphasized significant information generated lower retention results than no-added emphasis on the significant information. Our findings suggest that emphasis matters for both information comprehension and retention. Correctly placed emphasis can help listeners to understand the information better, but when we consider emphasis with pause, the information becomes less comprehensible.

Our study has several limitations. We generated the audio features using Azure's text-to-voice tool. We only used US male voice for our study. Using a female voice or a different accent might influence the result. Our goal was to evaluate adding emphasis and pauses that could be automated and added to our online editor as a suggestion. Better results may be obtained when individual writers add emphasis and pause for a text. However, to automatically identify where to add pause with Azure, more testing is needed. The effect of the emphasis and pause might differ in other settings. We cannot confirm the specific environment or audio devices the workers used while completing the HITs, but we can assume they were suitable. Moreover, while AMT workers reflect the broader public, results might change when actual patients evaluate the audio with endowed features.

## Acknowledgements

The research reported in this paper was supported by the National Library of Medicine of the National Institutes of Health under Award Number 2R01LM011975. The content is solely the responsibility of the authors and does not necessarily represent the official views of the National Institutes of Health. We also acknowledge the contribution of the AMT workers as they are the sole participants of this study.